\PassOptionsToPackage{table,xcdraw}{xcolor}
\documentclass{article} 
\usepackage{iclr2025_conference}
\usepackage{mathptmx}

\usepackage{microtype}
\usepackage{graphicx}
\usepackage{subcaption}
\usepackage{booktabs} 

\usepackage[hyphens]{url}

\usepackage{amsmath}
\usepackage{amssymb}
\usepackage{mathtools}
\usepackage{amsthm}

\theoremstyle{plain}

\theoremstyle{definition}

\theoremstyle{remark}

\usepackage[most]{tcolorbox}
\tcbset{
    promptbox/.style={
        colback=blue!5!white,    
        colframe=white,          
        boxrule=0mm,             
        sharp corners,           
        enhanced,                
        leftrule=0mm,            
        rightrule=0mm,           
        toprule=0mm,             
        bottomrule=0mm           
    }
}

\usepackage{array} 

\usepackage[textsize=tiny]{todonotes}
\setuptodonotes{inline} 

\newcommand{\ancgraph}{\mathcal{G}}
\newcommand{\predgraph}{\hat{\mathcal{G}}}
\newcommand{\probmat}{P}

\def\eqref#1{equation~\ref{#1}}

\def\1{\bm{1}}

\DeclareMathAlphabet{\mathsfit}{\encodingdefault}{\sfdefault}{m}{sl}
\SetMathAlphabet{\mathsfit}{bold}{\encodingdefault}{\sfdefault}{bx}{n}

\usepackage{url}

\definecolor{darkblue}{RGB}{0,0,139}
\usepackage[colorlinks=true, linkcolor=darkblue, citecolor=darkblue, urlcolor=darkblue]{hyperref}

\title{Large Language Models for Zero-shot \\ Inference of Causal Structures in Biology}

\author{
Izzy Newsham\textsuperscript{1}\thanks{Contributed equally.} \quad
Luka Kova\v{c}evi\'{c}\textsuperscript{1}\footnotemark[1] \quad
Richard Moulange\textsuperscript{1}\footnotemark[1] \\
\textbf{Nan Rosemary Ke}\textsuperscript{2, 3} \quad \textbf{Sach Mukherjee}\textsuperscript{4,1} \\
\textsuperscript{1}MRC Biostatistics Unit, University of Cambridge, Cambridge, UK \\
\textsuperscript{2}DeepMind, London, UK \\
\textsuperscript{3}Mila, Montreal, Canada \\
\textsuperscript{4}DZNE \& University of Bonn, Bonn, Germany \\
\texttt{\{izzy.newsham, sach.mukherjee\}@mrc-bsu.cam.ac.uk} \\
}

\preprintcopy

\begin{document}

\maketitle

\begin{abstract}
    Genes, proteins and other biological entities influence one another via causal molecular networks. Causal relationships in such networks are mediated by complex and diverse mechanisms, through latent variables, and are often specific to cellular context. It remains challenging to characterise such networks in practice. Here, we present a novel framework to evaluate large language models (LLMs) for zero-shot inference of causal relationships in biology. In particular, we systematically evaluate causal claims obtained from an LLM using real-world interventional data. This is done over one hundred variables and thousands of causal hypotheses. Furthermore, we consider several prompting and retrieval-augmentation strategies, including large, and potentially conflicting, collections of scientific articles. Our results show that with tailored augmentation and prompting, even relatively small LLMs can capture meaningful aspects of causal structure in biological systems. This supports the notion that LLMs could act as orchestration tools in biological discovery, by helping to distil current knowledge in ways amenable to downstream analysis. Our approach to assessing LLMs with respect to experimental data is relevant for a broad range of problems at the intersection of causal learning, LLMs and scientific discovery.
\end{abstract}

\section{Introduction}
Discovery in many scientific disciplines is complex
and often expensive and time-consuming. In biology, the large number of interacting components 
creates an enormous space of potential experiments to perform.
Experimental plans, as well as joint experimental-computational workflows, are usually informed by  existing literature (e.g. on a particular disease or biomolecular pathway), but this step 
is itself challenging. If large language models (LLMs) are capable of reducing even a small fraction of the work needed to elucidate biological mechanisms, or compress literature-derived knowledge into priors for \textit{in silico} models, they have the potential to significantly accelerate biological discovery and understanding of disease mechanisms.

To date, often without fine-tuning on biological data, LLMs have been applied in biochemistry as experimental orchestration tools \citep{m2024augmenting}, AI bioinformaticians \citep{ding2024automating}, collaborative multi-agent teams of AI scientists \citep{swanson2024virtual} or as a source for pre-trained embeddings for downstream prediction tasks \citep{chen2024genept}. However, to accelerate science, LLMs must be capable of 
elucidating  causal relationships, which is highly relevant in the 
context of studying molecular networks underpinning disease biology.

In experimental settings conducive to LLM-driven orchestration, researchers often face large experimental search spaces. Tasks may include 
elucidating the causal architecture underpinning disease phenotypes, 
identifying molecular pathways associated with a transcriptional signature, 
understanding the effects of perturbation of a specific gene target and more. LLMs that 
are capable of elucidating causal gene-gene relationships would be useful in collapsing the experimental space in the context of such tasks by, for example, 
identifying potential upstream regulators of a specific node, inferring
direct and indirect effects downstream of a given node and highlighting key regulatory pathways likely to be causally linked to specific nodes or processes in a given biological context.
However, for any such downstream application it would be essential to first 
understand how to evaluate whether LLMs are able to 
infer \textit{causal} relationships
between molecular entities.

Perturbation screens are fundamental  tools in biology that allow researchers to observe inherently causal relationships 
by intervening 
experimentally 
on specific
molecular entities. 
Contemporary gene perturbation experiments--- notably those using CRISPR-based gene editing tools---offer the possibility of carrying out interventions in a systematic fashion (spanning  potentially large numbers of intervention targets)
and to record changes brought about by such interventions in a global fashion.
These experimental approaches have developed rapidly and include a wide range 
of specific protocols \citep{jinek2012programmable, qi2013repurposing, dixit2016perturb, replogle2022mapping}.
Such experiments can be carried out in 
specific biological contexts (e.g. particular cell types and epigenetic backgrounds), offering a scalable way to explore causal relationships that may themselves be context-dependent.
Data from such experiments offer an opportunity to understand, in a systematic manner, whether  LLMs are capable of elucidating causal relationships between genes.

\textbf{Our contributions.} We leverage data from 
interventional 
experiments to generate a causal ground-truth to assess whether LLMs can infer gene regulatory networks in a zero-shot fashion, without prior exposure to the experimental data. 
Specifically, we focus on evaluating the ability of LLMs to identify directed gene--gene causal relationships, a key step towards their wider application in applied causal inference tasks in biology. 
Importantly, our experimental design tests LLM output with respect to empirical, interventional experiments, matching a real-world scientific  workflow. 
This paves the way to applying LLMs for the near-automated generation of context-specific priors for causal models in biology and helps to evaluate their potential in orchestrating experiments aimed at studying gene regulation. 
We offer the following contributions: \begin{enumerate}
    \item We develop a benchmarking approach for assessing the capabilities of LLMs in eludicating (directed) causal relationships in biology;
    \item We explore a range of retrieval-augmented strategies, including ways to specify biological or experimental context as well as providing guidance at the node or gene level;
    \item Finally, the LLM-based output is compared to standard knowledge-driven prior specification methods.
\end{enumerate}

By combining insights from causal structure learning, Perturb-seq datasets, and advanced text-mining tools like PubTator 3.0 \citep{wei2024pubtator} and the STRING database \citep{szklarczyk2023string}, this work aims to quantify the ability of current LLMs to infer causal gene--gene relationships and build a framework to help evaluate the role of LLMs in causal discovery within complex biological systems.

\section{Background}

We provide a brief introduction to single-cell perturbation screens in Section \ref{sec:scperturb}. We then introduce the basic terminology and notation from causal structure learning that will be used throughout this paper in Section \ref{sec:cslintro}. Finally, Section \ref{sec:relatedwork} describes related work.

\subsection{Single-cell perturbation screens} \label{sec:scperturb}

Advances in RNA sequencing technologies allow the measurement of thousands of molecular readouts in single cells, referred to as single-cell RNA sequencing (scRNA-seq). These advances underpin protocols such as Perturb-seq \citep{dixit2016perturb}, which combine large-scale  
interventions (e.g. gene knockouts) with measurement of  gene expression  at the single-cell level. Perturb-seq and related protocols allow researchers to explore 
causal effects at scale, providing rich datasets for studying gene networks and causal regulatory interactions. 

\subsection{Causal structure learning} \label{sec:cslintro} 

Inferring the existence of causal relationships is a fundamental question in science. Causal structure learning (CSL) methods 
\citep{heinze2018causal}
aim to identify these causal relationships from observational and interventional data. These relationships are usually represented by a causal graph \(\mathcal{G}=(V,E)\), which is composed of vertices \(V=[d]\) which are associated with variables $X_i, \, i \in V$.
The edge set $E$ comprises directed edges \(E=\{i \rightarrow j : X_i \in X_{\text{pa}(j)}\}\), i.e.
 an  edge \(i \rightarrow j\) indicates
that $X_i$ is a parent of $X_j$ in the casual graph. 

In this context, if there exists a directed path \(i \rightarrow \ldots \rightarrow j\), then node \(i\) is an ancestor of node \(j\), and conversely, node \(j\) is a descendant of node \(i\). 
Here, we use the term ancestral or indirect causal  graph to refer to a graph where each edge \(X_i \rightarrow X_j\) implies that \(X_i\) is an ancestor of \(X_j\), rather than necessarily a parent. We can also compute the transitive closure of $\mathcal{G}$, denoted by $\mathcal{G}^+ = (V,E^+)$, such that if there is a path between $i,j\in V$ then there is an edge $i\rightarrow j \in E^+$. Indirect edges are 
a relevant notion in analyzing experimental perturbations of the kind seen in genetic screens, since these capture total causal effects of perturbation on a given node \citep{hill2019}.

\subsection{Related work} \label{sec:relatedwork}

Previous research has demonstrated that LLMs are capable of extracting causal relationships between variables from short, well-structured sentences 
\citep{kiciman2023causal,nie2023moca,romanou2023crab,jin2024cladder}. However, these tasks typically lack the complexity, noise, and contradictions often found in scientific literature. 

\citet{martens2024enhancing} demonstrated that LLM-derived embeddings can improve the performance of generative models for perturbation prediction. While this highlights the utility of LLMs in representation learning for biology, it does not directly address their ability to perform causal inference tasks. By focusing specifically on causal retrieval and reasoning, our study fills this gap.

Other research in evaluation for CSL from Perturb-seq experiments has focused on evaluating CSL from Perturb-seq data directly \citep{chevalley2022causalbench, kovavcevic2024simulation}, or considered the perturbation prediction problem from a non-causal perspective \citep{wu2024perturbench, szalatabenchmark}. In \citet{chevalley2022causalbench} the authors apply a similar hypothesis testing procedure to generate ground truth causal graphs from Perturb-seq data, however, they focus on data-driven CSL methods. 

To evaluate LLM performance, 
in addition to testing against experimental data, 
we benchmark also against  methods
rooted in traditional knowledge-driven approaches. This includes the STRING database \citep{szklarczyk2023string}, a database that integrates known and predicted protein-protein interactions from multiple sources, including experimental data, computational predictions, and text mining. Unlike standard LLMs, STRING relies on curated biological networks and statistical association scores, making it highly structured and reliable for established interactions but potentially less adaptable to novel or ambiguous queries. Our work contrasts this classical knowledge-driven approach with the broader reasoning capabilities of LLMs.

\section{Methodology}

To evaluate the ability of LLMs to retrieve causal information about gene regulatory relationships, we compare the causal relationships identified by the LLM to those that can be inferred from Perturb-seq experiments. In Section \ref{sec:groundtruth}, we describe how the causal ground truth is constructed from perturbation data. Section \ref{sec:llm-prompting} briefly describes our prompting scheme, then Section \ref{sec:llmevals} defines the approach for evaluating the LLM-derived causal graph. 

\subsection{Constructing a causal ground truth} \label{sec:groundtruth}

\textbf{Problem setting.} For each Perturb-seq experiment with \(N\) cells we observe the pairs $\{\mathbf{x}^i, v_i\}_{i=1}^N$, where \(\mathbf{x}^i \in \mathbb{R}^{d}\) is the measured gene expression for $d$ genes and \(v_i\in V \: \cup \: \{I_0\}\) denotes the perturbation target, with \(I_0\) being the non-targeting perturbation (i.e. no gene is knocked out). Note that we assume each perturbation targets a single gene. We denote the univariate interventional distribution for the readout gene $j$
under intervention \(v_i\)  as \(\mathbf{X}^{v_i}_j\). 

\textbf{Hypothesis testing.} To identify the genes downstream of a given gene \(k\), we test the null hypothesis \(H_0:\mathbf{X}^{v_k}_j \sim \mathbf{X}^{I_0}_j\) (the intervened and unintervened univariate distributions for gene $j$, respectively \(\mathbf{X}^{v_k}_j\) and \(\mathbf{X}^{I_0}_j\), are identical) against the alternative hypothesis \(H_1: \mathbf{X}^{v_k}_j \not\sim \mathbf{X}^{I_0}_j\). The Mann--Whitney \textit{U} test \citep{mann1947test} is used for these comparisons, with the Benjamini--Hochberg correction \citep{benjamini1995controlling} to control for the false discovery rate across multiple hypothesis tests. A fixed significance level of \(\alpha = 0.05\) is used to determine significant differences.

This procedure identifies a set of differentially expressed genes\footnote{Differentially expressed genes are those that are significantly different between two conditions. In this case, the two conditions are before and after a targeted (CRISPRi) intervention.} for gene $k$ denoted by $\Delta_k$, where each $j\in\Delta_k$  represents a gene that significantly changes following an intervention on gene $k$. Formally, we define: \begin{equation}\Delta_k = \{j \mid p_j^k < \alpha\},\end{equation} where $p_j^k$ is the corrected p-value from the hypothesis test comparing the distributions of $X^{v_k}_j$ and $X^{I_0}_j$.

\begin{figure}[!ht]
\centering
\begin{tikzpicture}[node distance=0.75cm,>=stealth, thick]
    \fill[blue!5!white] (-1.5,-0.5) rectangle (1.5,0.5); 
    \draw[rounded corners] (-1.5,-0.5) rectangle (1.5,0.5);
    \node[below] at (1,-0.5) {\(\Delta_k\)};

    \node[circle, draw, fill=white] (i) at (-1,0) {$i$};
    \node[circle, draw, fill=white] (j) at (1,0) {$j$};
    \node[circle, draw, fill=white] (X) at (0,1.5) {$k$};

    \node[circle, draw] (i) at (-1,0) {$i$};
    \node[circle, draw] (j) at (1,0) {$j$};
    \node[circle, draw] (X) at (0,1.5) {$k$};
    
    \node at (-0.125,0) {$\cdot$};
    \node at (0,0) {$\cdot$};
    \node at (0.125,0) {$\cdot$};

    \draw[->] (X) -- (i);
    \draw[->] (X) -- (j);
\end{tikzpicture}
\caption{Directed edges are drawn between the perturbed gene $k$ and the set of genes $\Delta_k=\{i, \ldots, j\}$ that change significantly under experimental intervention on $k$.}
\label{fig:diff-gene-graph}
\end{figure}
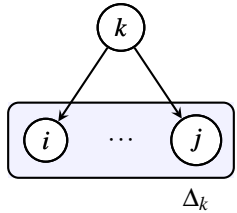

As shown in Figure \ref{fig:diff-gene-graph}, for each $j\in\Delta_k$ we draw an ancestral edge $k \rightarrow j$, signifying that gene $k$ causally influences gene $j$. Repeating this procedure across all $k \in \{1, \ldots, d\}$ yields our baseline ancestral graph $\ancgraph$. Each directed edge in $\ancgraph$ is interpreted as causal but ancestral and possibly indirect, as discussed in Section \ref{sec:cslintro}.

\subsection{LLM prompting for ancestral causal graphs} \label{sec:llm-prompting}

The LLM is prompted to predict the probability of a causal relationship between each pair of genes $i,j \in V$, which we call the \textit{query} genes. For example\footnote{Details of our prompting strategy are given in Appendix \ref{app:prompt_details}.}:

\begin{tcolorbox}[promptbox]
\verb|User|: Please quantify the extent to which gene $i$ has a causal effect on gene $j$. Return your answer as a two decimal place probability between 0 and 1.
\end{tcolorbox}

The LLM output is parsed automatically to retrieve the predicted probability. If no probability is present, it is set to 0, however, this affected only $<0.33\%$ across all prompts in our experiments. 

We repeat this process for all possible pairs of $d$ genes to obtain a matrix of probabilities $\probmat\in[0,1]^{d\times d}$ where each entry $P[i,j]=p_{i\rightarrow j}$ represents the probability of an edge $i \rightarrow j$. In general, this matrix is not symmetric since the probability of an edge 
$i\rightarrow j$ may differ from $j\rightarrow i$.
For performance metrics that require binary calls on edges, including those used in this work, the matrix $P$ is thresholded at $\gamma$ to yield 
a predicted graph $\predgraph_\gamma$ or its transitive closure $\predgraph_\gamma^+$.
Crucially, after each prompt, the LLM is reinstatiated to prevent biasing the output.

\subsection{Evaluation} \label{sec:llmevals}

Given a probability matrix $P$ obtained by repeatedly prompting an LLM and a ground truth $\ancgraph$, 
we evaluate  performance using Area Under the Receiver Operating Characteristic curve (AUROC).
We calculate the AUROC over all non-diagonal items in $P$, where the binary labels are given by the corresponding items in the adjacency matrix of $\ancgraph$.
In Appendix \ref{app:transitive}, we also present results from considering transitive closures of the ground truth and predicted graphs.

\section{Experiments} \label{sec:exp}

\subsection{Dataset} \label{sec:dataset}

The Perturb-seq dataset generated by \citet{replogle2022mapping} contains data from more than 2.5 million human cells on cell-lines\footnote{Cell-lines are a class of immortalised  cells commonly used in biology as tractable laboratory models.} K562 and RPE1. This dataset has been used in several causal modelling papers to study the performance of causal models in perturbation prediction \citep{chevalley2022causalbench}.

Since a causal graph with no self-loops and $d$ nodes requires $n(n-1)$ LLM queries
to facilitate exploration of various strategies we consider only the 100 most commonly referenced cancer relevant genes in constructing the causal graph. The process of selecting these genes is explained in Appendix \ref{app:gene_data}.

The data is filtered to exclude lowly expressed genes and cells, and $z$-normalised with respect to the unperturbed (control) samples,
\begin{align}
    \tilde{\mathbf{x}}^i=\frac{\mathbf{x}^i-\mu_0}{\sigma_0},
\end{align}
where $\mu_0$ is the mean and $\sigma_0$ is the standard deviation of $\mathbf{X}^{I_0}$. The filtered and $z$-normalised dataset $\{\tilde{\mathbf{x}}^i, v_i\}^N_{i=1}$ is then used to generate $\ancgraph$ as detailed in Section \ref{sec:groundtruth}.

\subsection{LLM prompting settings} \label{sec:prompting-strategies}

For all our experiments, we use \verb|Gemma2-9B-it| \citep{team2024gemma}, from the state-of-the-art small-model family at time of publication. This is because it is open-source, so we could freely test different experimental settings in the early project exploration stage, and relatively small, so that it required only one GPU to run. Given that some of our experiments---those that introduce long additional literature contexts---took up to 32h to run, we were unable to straightforwardly use larger models and these restrictions are relevant also to prospective scientific use-cases.

In addition to evaluating the performance of LLM-based retrieval without prior or contextual information, we consider additions to our original prompt that allow us to condition the LLM's output on the cellular context, measurement modality, gene function and experimental protocol. We reason that providing additional information about the experimental and biological context could help the LLM reach an embedding space more closely related to the experimental context. 

The full list of LLM prompts used for both the experiments for inferring causal direction and full causal graphs can be found in Table \ref{tab:prompt-context}.

\begin{table*}[h!]
\centering
\small
\rowcolors{2}{gray!10}{white} 
\begin{tabular}{p{0.15\linewidth}p{0.35\linewidth}p{0.15\linewidth}p{0.2\linewidth}}
\toprule
\textbf{Setting}             & \textbf{Description}                    & \textbf{Context Type} & \textbf{Inference task} \\ \midrule
\texttt{naive} & Default prompt. & None & Causal direction (\ref{sec:causal-direction}) \& graph (\ref{sec:causal-graph}) \\
\texttt{cancer}  & Default prompt with context about the type of cancer that the cells come from. & Experimental & Causal direction (\ref{sec:causal-direction}) \& graph (\ref{sec:causal-graph}) \\
\texttt{gene-desc}  & Default prompt with independent information about each of the query genes. & Query genes & Causal direction (\ref{sec:causal-direction}) \\
\texttt{literature}  & Default prompt with extracts from the literature describing the causal relationship between the query genes. & Query genes & Causal direction (\ref{sec:causal-direction}) \& graph (\ref{sec:causal-graph}) \\
\midrule
\texttt{false}  & Default prompt with a statement denying the true causal relationship between the query genes. & Query genes & Causal direction (\ref{sec:causal-direction}) \\
\texttt{contradict}  & Default prompt with multiple contradictory statements about the causal relationship between the query genes. & Query genes & Causal direction (\ref{sec:causal-direction}) \\
\midrule 
\texttt{mRNA}                       & Default prompt with context about the type of measurement being taken, which in this case is mRNA abundance or gene expression. & Experimental & Causal graph (\ref{sec:causal-graph}) \\
\texttt{evidence}              & Default prompt with encouragement for the LLM to consider various types of evidence present in the literature. & Experimental & Causal graph (\ref{sec:causal-graph}) \\
\texttt{cancer} \newline \hspace{1em} + \texttt{mRNA}               & Combination of the above prompt settings. & Experimental & Causal graph (\ref{sec:causal-graph}) \\
\texttt{cancer} \newline \hspace{1em} + \texttt{mRNA} \newline \hspace{1em} + \texttt{evidence} & Combination of the above prompt settings. & Experimental & Causal graph (\ref{sec:causal-graph}) \\
\texttt{cancer} \newline \hspace{1em} + \texttt{mRNA} \newline \hspace{1em} + \texttt{experiment} & Combination of above prompt settings with additional context about the type of CRISPRi experiment carried out. & Experimental & Causal graph (\ref{sec:causal-graph}) \\
\bottomrule
\end{tabular}
\caption{Prompt settings with a description of what they should contain and the task where they are used. The table also includes the type of information that each setting targets whether regarding the query genes or the experimental setting. Examples of each prompt setting can be found in Appendix \ref{app:prompt_details}.} 
\label{tab:prompt-context}
\end{table*}

\subsection{Inferring causal direction} \label{sec:causal-direction}

\begin{figure*}[t]
  \centering
  \includegraphics[width=\textwidth]{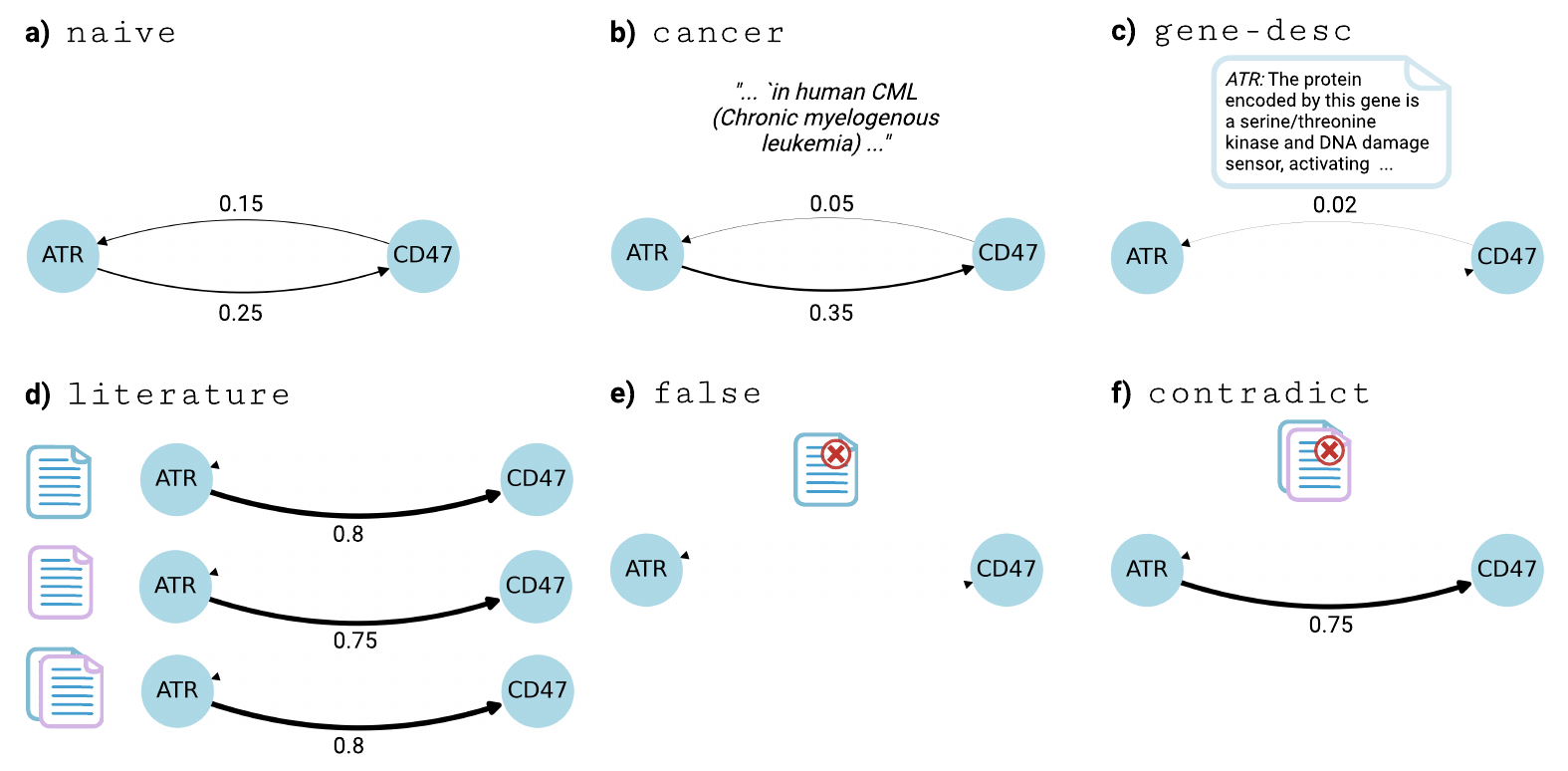}
  \vspace{-20pt}
  \caption{Outputs for inferring causal direction with different prompt contexts, for the example gene pair ATR and CD47.}
  \label{fig:proof_of_concept}
\end{figure*}

As an initial illustrative example, we investigate the LLM's ability to infer the causal direction between the example gene pair ATR and CD47. For this pair, $\ancgraph$ includes the causal edge ATR $\rightarrow$ CD47, but not CD47 $\rightarrow$ ATR (i.e. the experiments support the former but not the latter). Below, we outline various strategies using these particular genes to exemplify the approaches; in the sequel 
the approaches sketched below are applied systematically over all gene pairs.

\textbf{Naive inference.}  With no contextual information, the LLM returns $p_{\text{ATR}\rightarrow\text{CD47}}=0.25$ and $p_{\text{CD47}\rightarrow\text{ATR}}=0.15$, as visualised in \ref{fig:proof_of_concept}\textbf{a}. This simple two gene problem provides an example where the LLM is capable of distinguishing causal direction (here, given that there is a causal relationship) with only the names of the two genes (systematic results follow below).

\textbf{Cancer contextual information.} 
Next, we explore how conditioning the LLM on the 
relevant cancer type affects its causal assessment. We add \textit{`in human CML (Chronic myelogenous leukemia)'} to the prompt, which specifies the disease type of the K562 cells on which the experiments were performed. This improves the LLM's predicted scores yielding $p_{\text{ATR}\rightarrow\text{CD47}} = 0.35$ and $p_{\text{CD47}\rightarrow\text{ATR}} = 0.05$, as visualised in \ref{fig:proof_of_concept}\textbf{b}.

\textbf{Gene-wise contextual information.} We consider several approaches to conditioning on gene-relevant information. First, we consider general independent descriptions of ATR and CD47, provided by RefSeq \citep{o2016reference}, which leads to significantly lower predicted probability scores for both causal directions ($p_{\text{ATR}\rightarrow\text{CD47}} = 0$, $p_{\text{CD47}\rightarrow\text{ATR}} = 0.02$), as shown in Figure \ref{fig:proof_of_concept}\textbf{c}.

Now we consider relevant context from the literature comprised of passages\footnote{Full gene context passages can be found in Table \ref{tab:prompt_details_gene_specific}.} where the causal link between ATR and CD47 is directly referenced. The probabilities with this additional context are $p_{\text{ATR}\rightarrow\text{CD47}} \geq 0.75$ and $p_{\text{CD47}\rightarrow\text{ATR}} = 0$. This is visualised in \ref{fig:proof_of_concept}\textbf{d} and shows that provided sufficient information directly, LLMs can correctly predict causal direction. It has been shown previously by \citet{jin2024cladder} that LLMs are capable of this kind of causal reasoning. This underlines again that LLM performance on causal tasks is dependent on the provided evidence.

\textbf{Contradictory contextual information.} To explore this further, we examine the effect of incorrect sentences and contradictory sentences. As expected, the incorrect sentence describing no causal link from ATR to CD47 results in a predicted score of 0 in both directions (Figure \ref{fig:proof_of_concept}\textbf{e}). When provided with contradictory sentences, the LLM is still able to predict the correct causal edge with $p_{\text{ATR}\rightarrow\text{CD47}} = 0.75$ and $p_{\text{CD47}\rightarrow\text{ATR}} = 0$ (Figure \ref{fig:proof_of_concept}\textbf{f}).

\subsection{Inferring a causal graph} \label{sec:causal-graph}

Having looked at an example where LLMs were capable of distinguishing causal direction (in the specific example, given that there was a causal relationship), we now turn to the more general and challenging problem of inferring a full causal graph for the genes specified in Section \ref{sec:dataset}.

\begin{figure*}[ht!]
\centering
\includegraphics[width=\textwidth]{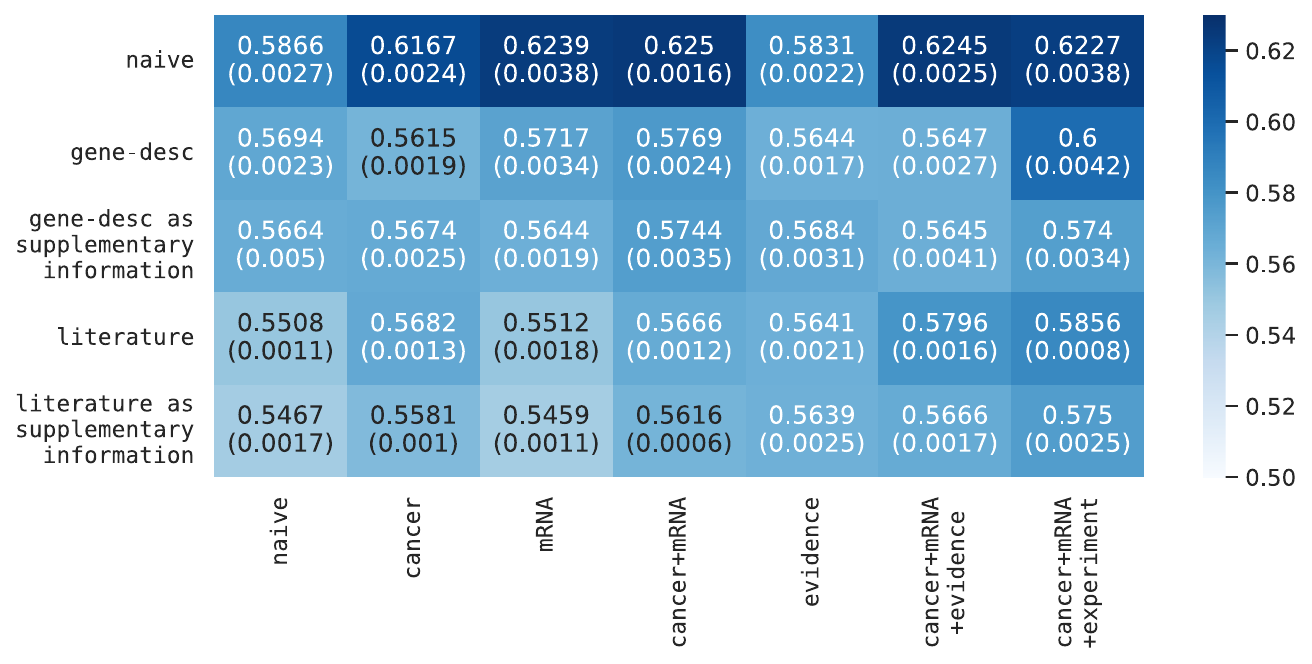}
\vspace{-20pt}
\caption{Results on Gemma2 for all combinations of prompt variants (different contexts along each column, different gene-specific information along each row). The results are shown as the mean AUROC over 10 repetitions, with the standard error given in brackets.}
\label{fig:gemma2_results_matrix}
\end{figure*}

Figure \ref{fig:gemma2_results_matrix} shows the results for \verb|Gemma2| with varying degrees of gene-wise contextual information on the y-axis and experimental contextual information on the x-axis, as described in Table \ref{tab:prompt-context}. Each cell shows the mean AUROC score for this level of context with the standard error in parentheses. The distributions of the AUROC scores are visualised in Figure \ref{fig:gemma2_results_boxplots}.

The best performance is an AUROC $=0.625$, via prompting with the experimental context of \texttt{cancer} + \texttt{mRNA} with no gene-specific information. Providing \texttt{literature} evidence near-uniformly reduces the LLM performance, especially when there are additional caveats that the literature may not generalise to the specific context of the experiments in question. We hypothesise that this is because this further prompting introduces additional uncertainty and leads the LLM to become less likely to commit to high-probability outputs. When not including gene-specific information, \texttt{cancer} +\texttt{mRNA} yield predictions with AUROC $>0.6$. Including additional details on experimental protocol generally improves performance slightly.

Interestingly, repeating these experiments with chain-of-thought (CoT) reasoning \citep{wei2022chain} does not improve AUROC scores. Figure \ref{fig:gemma2_results_cot_simple} shows this for a simple variant of CoT and Figure \ref{fig:gemma2_results_cot_guided} for a more detailed, guided variant of CoT, which shows that CoT decreases the performance in many cases. Again, we suspect this is because chain-of-thought reasoning encourages the LLM to be less confident for or against a causal relationship.

\textbf{Gene-wise contextual information.} At least in the present setting of a small LLM 
in the context of 
specific ground-truth experiments, gene descriptions and literature information appear to not be useful for predicting causal regulatory relationships observed in the experiments. We believe that this may be due to the lack of literature available for mechanisms impacted by CRISPRi interventions used to generate the \citet{replogle2022mapping} dataset. The prompts used for the final row of results in Figure \ref{fig:gemma2_results_matrix} emphasise that the gene-specific contextual information should only be used as supplementary information and yet this worsens performance across prompt settings.

To investigate why gene-wise contextual information worsens performance, we quantify the correlation between $\ancgraph$ and the literature evidence: there is no correlation between $\ancgraph$ and the literature (Boschloo's one-sided exact test: $p=0.2075$; \citet{boschloo1970raised}). That is, the presence of literature evidence for a gene pair is not significantly associated with whether that gene pair is causal, as described in Appendix \ref{app:gene_data}. This result, 
over a large number of causal relationships,
underlines the 
limitations of relying on traditional literature mining approaches in the context of elucidating causal relationships at large-scale in specific biological contexts. 

We also find that the gene descriptions, even when provided as supplementary information, strongly influence the LLM's predictions. When examining the outputs of the LLM with guided CoT, we find that the model heavily focuses on the distinct functions of the genes and how their pathways are not directly interconnected. For example, one of the outputs for the causal gene pair ATR and CD47 states: ``There is no readily apparent direct relationship between ATR's role in DNA damage response and repair, and CD47's function in cell adhesion, calcium signaling, and thrombospondin binding" and ``A lack of known direct interactions or regulatory pathways linking ATR to CD47 suggests a low probability of a causal effect".

\textbf{Comparison with a knowledge-driven baseline.} Next, we investigate how the predictions from \verb|Gemma2| compare to a method that does not utilise LLMs but also draws from existing literature-based knowledge on gene interactions and associations. We obtain the scores given by the STRING database (as described in Appendix \ref{app:gene_data}), which integrates multiple gene interaction and association databases and literature-derived associations, and evaluate them against an undirected version of $\ancgraph$. This results in an AUROC of 0.460, demonstrating that the predictions by \verb|Gemma2| outperform the predictions obtained from STRING for this task regardless of the prompt setting used. It is important to note however, that the STRING database provides symmetric association scores, which are not intended to provide causal information.

\bigskip


\section{Discussion}

In conclusion, we find that even a small LLM is capable of inferring causal gene--gene relationships better than random chance and in an entirely zero-shot manner, with no input experimental data.  \verb|Gemma2| performs considerably better than a database-based zero-shot baseline based on the STRING database. While the highest AUROC of 0.625 is not large in absolute terms, this is an entirely automated, zero-shot method, which requires no biological knowledge nor any empirical data for the task at hand. The LLM output can be straightforwardly integrated as a prior into any downstream causal structure learning task. Due to the hyperexponential space of possible causal graphs, even a weak prior that could steer data driven methods has the potential to signficantly narrow this space. 
\par
Our results highlight the importance of context-specific information. Potentially, more sophisticated retrieval-augmented generation strategies---with careful constraining of the associated context---may improve performance. We found that providing the experimental context improved performance yet gene-specific information did not, suggesting that the latter lacked specificity to enable accurate inference with respect to the target ground truth of interest. Indeed, this underlines the need for novel approaches, that go beyond classical literature mining, in the context of elucidating causal relationships in biology. Future work could search explicitly for experimentally-relevant literature---in this case, relating to CML cells and mRNA measurements---rather than any biological or experimental context with associated gene measurement.
\par
Although we spent significant effort on prompt engineering, we nevertheless expect that it might be possible to further improve results by providing longer context and optimising the prompts further. This is particularly the case for the chain-of-thought reasoning. Moreover, larger `vanilla' LLMs or RL-imbued reasoning-style LLMs (such as o1; \citet{jaech2024openai}) may perform better. As with all recent LLM advances, this work represents a lower-bound on what may be achievable and 
it will be interesting to see whether future work and more advanced LLMs can provide more accurate zero-shot CSL priors and better quantify causal gene--gene relationships at scale.




\bigskip

\textbf{Acknowledgements:} This work was partly supported by the UK Medical Research Council (MC\_UU\_00040/5 \& MC\_UU\_00002/17), the Helmholtz Association AI Project ``UNITY" and the National Institute for Health Research (Cambridge Biomedical Research Centre at the Cambridge University Hospitals NHS Foundation Trust).

\bibliography{iclr2025_conference}
\bibliographystyle{iclr2025_conference}

\appendix
\section*{Appendix}

\renewcommand{\thefigure}{A.\arabic{figure}}
\setcounter{figure}{0}
\renewcommand{\thetable}{A.\arabic{table}}
\setcounter{table}{0}

\section{Prompting details}
\label{app:prompt_details}

For each pair of genes---in the 9,900 non-identical pairs constructed from the 100 genes pre-selected for their relevance to cancer (see Appendix \ref{app:gene_data})---we used the following default prompt. Here, we have provided the example with genes ATR and CD47:
\vspace{-3pt}
\begin{tcolorbox}[promptbox]
\verb|User|: You are a professional biologist who is an expert in assessing whether one gene has a causal effect or not on another. You reply concisely and always follow instructions exactly. Below, you will be asked to quantify the level at which one gene has a causal effect on another by returning a 2 decimal place number between 0 and 1, where 0 indicates the first gene has no causal effect on the second and 1 indicates the first gene has a very strong causal effect on the second. Here is an example:\\
{[}Begin example{]}\\
Q: Please quantify the extent to which [Gene A] has a causal effect on [Gene B]. Return your answer as a 2 decimal place probability between 0 and 1.\\
LLM output: Probability = {the 2dp probability}\\
{[}End example{]}\\
IMPORTANT: DO NOT REPLY IN ANY OTHER WAY. YOUR REPLY MUST END WITH ``Probability = " AND THEN A 2 DECIMAL PLACE PROBABILITY. Remember, you are a highly accurate expert biologist who answers concisely, follows instructions and returns only 2dp probabilities as answers.\\
Q: Please quantify the extent to which ATR has a causal effect on CD47. Return your answer as a 2 decimal place probability between 0 and 1.\\
Probability =
\end{tcolorbox}

We constructed 105 different variants of this prompt by considering three different types of augmentation: the \textit{experimental context} that underpins the ground truth data we used to evaluate the LLM inferences, \textit{specific information on the gene--gene pair} in question and requests for \textit{chain-of-thought reasoning}. After drafting the prompts ourself, we used GPT-4o \cite{hurst2024gpt} to improve the prompts.
\par
We investigated providing six additional experimental contexts to better guide the LLM to the correct biologically-relevant part of its embedding space. These included clarifying that the prediction should be based on a specific type of cancer, that the causal relationship was identified using mRNA gene expression data, and sometimes additional experimental or biological details about causal gene--gene experiments. See Table \ref{tab:prompt_details_context} for the specific modifications we made to the prompts. We also provided four different types of gene-specific information (see Table \ref{tab:prompt_details_gene_specific}). The first two straightforwardly drew on gene descriptions from RefSeq, with or without an additional note clarifying that this was only supplementary information. The third and fourth types provided literature evidence using the PubTator 3.0 system, again with or without a clarifying note. Finally, we experimented with two chain-of-thought reasoning extensions, which are described in Table \ref{tab:prompt_details_CoT}. The maximum number of new tokens was set differently for different chain of thought variants: 500 for guided, 200 for simple and 10 for no chain-of-thought.
\par
For around 40\% of the gene pairs, no PubTator associations were found. For such cases---for example HMGB1 and RUNX1---the PubTator part of the prompt was replaced with
\begin{tcolorbox}[promptbox]
A search of PubTator for research articles relating to HMGB1 and RUNX1 did not yield relevant results. However, the absence of findings in this search does not rule out the possibility of an association, as data may exist in other resources or contexts not captured by PubTator.
\end{tcolorbox}
to clarify. If more than 100 PubTator associations were found, only the first 100 were given due to memory limitations.

\section{Gene data}
\label{app:gene_data}
\textbf{Cancer genes.} We used a text-mining procedure to obtain 100 well-known cancer genes. Specifically, we used the Entrez Direct tool \citep{kans2013entrez} and Pangaea \citep{pirvan2020pangaea} to parse abstracts from PubMed articles using the search term ``cancer". From these abstracts, we identified the 100 most commonly referenced genes that were both perturbed and measured in the Replogle dataset.

\textbf{Gene descriptions.}
We used the Entrez Direct tool \citep{kans2013entrez} to retrieve the Entrez gene summary for each gene.

\textbf{Literature evidence}
For each gene pair and each of the following relation types: ``associate", ``interact", ``positive\_correlate", ``negative\_correlate",
we queried the PubTator 3.0 search API \citep{wei2024pubtator} and extracted the resulting text to produce the literature evidence data. We ran a one-sided Boscholoo's exact test on the literature evidence data to investigate whether the presence of literature evidence was significantly associated with whether a particular gene--gene pair had a causal relationship in either direction. Specifically, we split the gene pairs into two groups: gene pairs with at least one sentence in the literature evidence, and gene pairs with no such sentences. We constructed a contingency table (see Table \ref{tab:pubtator_fishers}): A one-sided Boschloo's exact test gives a nonsignificant $p$-value of 0.2075, which suggests the presence of literature evidence is not significantly associated with whether the gene--gene pair is causal.

\textbf{STRING scores.} We downloaded the STRING database \citep{szklarczyk2023string} for \textit{Homo sapien} and extracted the combined scores for each gene--gene pair. We constructed a \(d \times d\) matrix from the scores (which is symmetric since STRING scores are symmetric), with 0-entry where no STRING score was found.

\section{Evaluation under transitive closures} \label{app:transitive}

In the main text, we have assumed that the LLM infers \textit{ancestral} causal relationships, since our prompt does not specify that direct causal edges are needed but rather whether gene A has a causal effect on gene B. This causal effect could occur due to a direct causal relationship (i.e. $A \rightarrow B$) or through an indirect path (i.e. $A \rightarrow C \rightarrow B$).

Thus, the LLM's predictions might represent direct edges or even a mix of ancestral and direct edges, hence to assess against an indirect ground truth it may be helpful to consider the transitive closure of the LLM-derived graph. We therefore considered the transitive closure of $\predgraph_\gamma$, denoted by $\predgraph_\gamma^+$, and compared this with the ground truth $\mathcal{G}$. We also considered comparing $\predgraph_\gamma^+$ to the transitive closure of the ground truth, $\ancgraph^+$.

\begin{figure*}[!ht]
\centering
\includegraphics[width=0.85\textwidth]{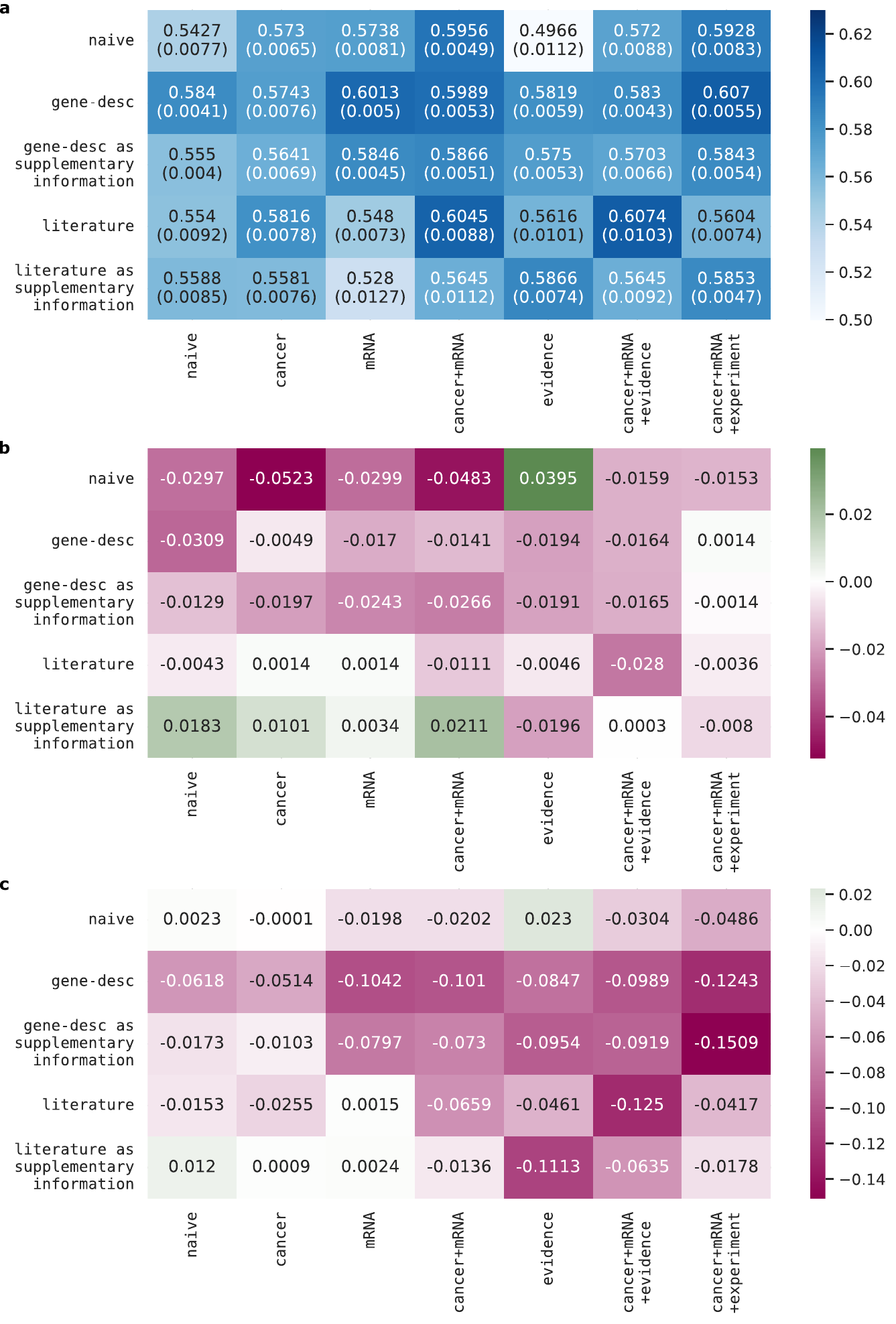}
\caption{Results on Gemma2 for all combinations of prompt variants when computing the transitive closure over the predictions. 
        \textbf{a}) The results for no CoT, showing the mean AUROC over 10 repetitions, with the standard error given in brackets.
        \textbf{b}) The results for simple CoT, compared to the results using no CoT.
        \textbf{c}) The results for guided CoT, compared to the results using no CoT.
        }
\label{fig:gemma2_results_transitive_closure}
\end{figure*}

\begin{figure*}[!ht]
\centering
\includegraphics[width=0.85\textwidth]{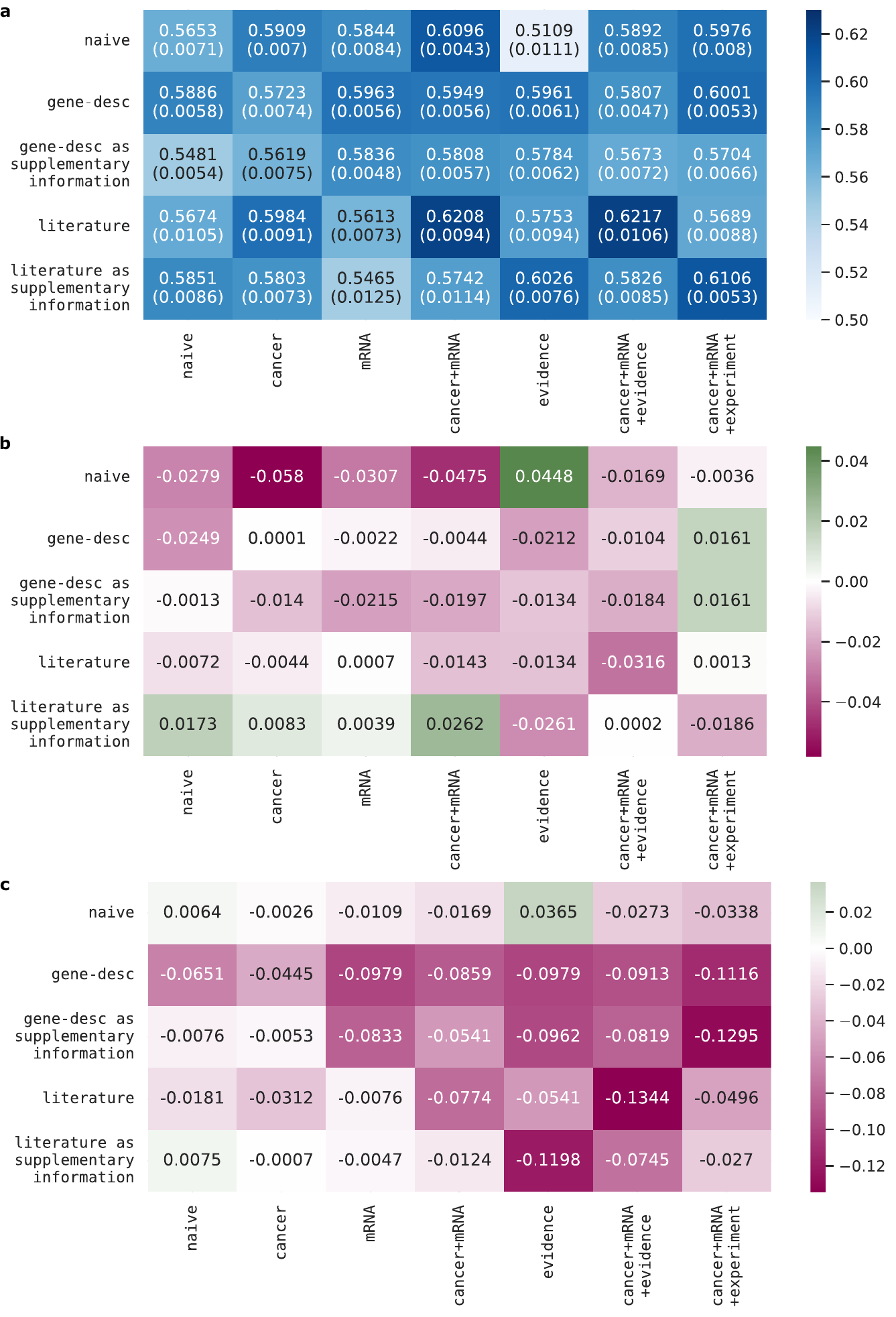}
\caption{Results on Gemma2 for all combinations of prompt variants when computing the transitive closure over both the predictions and the ground truth. 
        \textbf{a}) The results for no CoT, showing the mean AUROC over 10 repetitions, with the standard error given in brackets.
        \textbf{b}) The results for simple CoT, compared to the results using no CoT.
        \textbf{c}) The results for guided CoT, compared to the results using no CoT.
        }
\label{fig:gemma2_results_transitive_closure_over_preds_and_gt}
\end{figure*}

We first compare $\predgraph^+_\gamma$ to the ground truth $\ancgraph$. The results are shown in Figure \ref{fig:gemma2_results_transitive_closure}\textbf{a}. This shows the AUROCs are generally lower than before, but here \texttt{gene-desc} and \texttt{literature} improve performance in some cases, contrasting with our previous results. The best performing prompt variant is \texttt{literature} and \texttt{cancer}+\texttt{mRNA}+\texttt{evidence} with an AUROC of 0.6074.
This could imply that literature evidence helps \verb|Gemma2| to identify direct causal edges, but is less helpful for identifying the ancestral causal edges.
We also compare $\predgraph_\gamma^+$ to the transitive closure of the ground truth, $\ancgraph^+$, and obtain similar results, as shown in Figure \ref{fig:gemma2_results_transitive_closure_over_preds_and_gt}\textbf{a}.

Figures \ref{fig:gemma2_results_transitive_closure}\textbf{b} and \textbf{c} (and \ref{fig:gemma2_results_transitive_closure_over_preds_and_gt}\textbf{b} and \textbf{c}) show how the AUCs change with simple and guided CoT, respectively. As before, the AUCs do not improve with CoT and in some cases the guided CoT decreases performance by up to 0.15.

\section{Training details}
Each inference experiment was completed on a single NVIDIA A100-SXM-80GB GPU, using resources provided by the Cambridge Service for Data Driven Discovery (CSD3) operated by the University of Cambridge Research Computing Service, provided by Dell EMC and Intel using Tier-2 funding from the Engineering and Physical Sciences Research Council (capital grant EP/T022159/1), and DiRAC funding from the Science and Technology Facilities Council. In particular, each GPU node contained four such GPUs, with two AMD EPYC 7763 64-Core 1.8GHz Processors and dual-rail Mellanox HDR200 InfiniBand interconnect.

\begin{table}[h!]
\caption{Prompts with additional context}
\centering
\small
\renewcommand{\arraystretch}{1.5} 

\begin{tabular}{p{4cm}p{11cm}} 
\textbf{Context} & \textbf{Prompt addition} \\ \toprule
+cancer                   &  \dots \ [Gene A] has a causal effect on [Gene B]\textbf{ in human CML (Chronic myelogenous leukemia)} \ \dots \newline
\dots \ ATR has a causal effect on CD47\textbf{ in human CML (Chronic myelogenous leukemia)}\ \dots \\ 
+mRNA                     &  \dots \ [Gene A] has a causal effect on [Gene B]\textbf{ in the context of gene expression (mRNA measurements)}\ \dots \newline
\dots \ ATR has a causal effect on CD47\textbf{ in the context of gene expression (mRNA measurements)}\ \dots \\ 
+cancer+mRNA              &  \dots \ [Gene A] has a causal effect on [Gene B]\textbf{ in human CML (Chronic myelogenous leukemia) in the context of gene expression (mRNA measurements)}\ \dots \newline
\dots \ ATR has a causal effect on CD47\textbf{ in human CML (Chronic myelogenous leukemia) in the context of gene expression (mRNA measurements)}\ \dots \\ 
+extra detail             & \dots \ a very strong causal effect on the second. \textbf{Use known scientific evidence, including experimental findings, gene pathway involvement, and established literature, to determine the causal effect. Prioritise experimental findings, such as knockout studies and mechanistic insights from direct gene interactions. Assess the strength of the evidence and consider the possibility of unknown factors. If causal relationships between two genes are unclear or lack sufficient evidence, return a probability reflective of this uncertainty. } Here is an example:\ \dots \\ 
+cancer+mRNA+extra detail & \dots \ a very strong causal effect on the second. \textbf{Use known scientific evidence, including experimental findings, gene pathway involvement, and established literature, to determine the causal effect. Prioritise experimental findings, such as knockout studies and mechanistic insights from direct gene interactions. Assess the strength of the evidence and consider the possibility of unknown factors. If causal relationships between two genes are unclear or lack sufficient evidence, return a probability reflective of this uncertainty.} Here is an example: \newline 
\dots \ [Gene A] has a causal effect on [Gene B]\textbf{ in human CML (Chronic myelogenous leukemia) in the context of gene expression (mRNA measurements)}\ \dots \newline
\dots \ ATR has a causal effect on CD47\textbf{ in human CML (Chronic myelogenous leukemia) in the context of gene expression (mRNA measurements)}\ \dots \\ 
+Perturb-seq details      & ..a very strong causal effect on the second. \textbf{The evaluation is based on single-cell data from a Perturb-seq experiment conducted in K562 cells, a chronic myeloid leukemia (CML) cell line. In this experiment, gene perturbations are systematically introduced to assess their effects on the expression levels of other genes at single-cell resolution. Your task is to estimate the extent to which Gene B is affected by the knockdown of Gene A in this specific experimental context (without seeing the raw experimental data). The output probability should estimate the causal relationship inferred from the Perturb-seq data.} Here is an example:\newline 
\dots \ [Gene A] has a causal effect on [Gene B]\textbf{ in the context of this Perturb-seq experiment}\ \dots \newline
\dots \ ATR has a causal effect on CD47\textbf{ in the context of this Perturb-seq experiment}\ \dots \\ \bottomrule
\end{tabular}
\label{tab:prompt_details_context}
\end{table}

\begin{table}[ht]
\caption{Prompts with gene-specific information, for the gene pair ATR and CD47}
\centering
\small
\renewcommand{\arraystretch}{1.5} 
\begin{tabular}{p{2.5cm}p{11.5cm}} 
\textbf{Gene-specific \newline information} & \textbf{Prompt addition} \\ \toprule
Gene descriptions & \dots \ follows instructions and returns only 2dp probabilities as answers.\newline
\textbf{You may use the following contextual information:\newline
Description of ATR: The protein encoded by this gene is a serine/threonine kinase and DNA damage sensor, activating cell cycle checkpoint signaling upon DNA stress. The encoded protein can phosphorylate and activate several proteins involved in the inhibition of DNA replication and mitosis, and can promote DNA repair, recombination, and apoptosis. This protein is also important for fragile site stability and centrosome duplication. Defects in this gene are a cause of Seckel syndrome 1. [provided by RefSeq, Aug 2017]\newline
Description of CD47: This gene encodes a membrane protein, which is involved in the increase in intracellular calcium concentration that occurs upon cell adhesion to extracellular matrix. The encoded protein is also a receptor for the C-terminal cell binding domain of thrombospondin, and it may play a role in membrane transport and signal transduction. This gene has broad tissue distribution, and is reduced in expression on Rh erythrocytes. Alternatively spliced transcript variants have been found for this gene. [provided by RefSeq, Jul 2010]}\newline
Q: Please quantify the extent \ \dots \\
Gene descriptions as supplementary information & \dots \ follows instructions and returns only 2dp probabilities as answers.\newline
\textbf{You may use the following contextual information:\newline
Gene Descriptions: These provide independent functional information about each gene. Treat them as supplementary context, not definitive evidence of causality. While contrasting or compatible functions may inform your reasoning, rely primarily on broader biological knowledge, including regulatory pathways, experimental evidence, and known gene interactions.\newline
Description of ATR: \ \dots \newline
Description of CD47: \ \dots \newline
Note: The descriptions above are intended to provide auxiliary context about each gene's independent functions. They may suggest plausible functional relationships or incompatibilities but are not expected to directly indicate causality. Use them as supplementary information alongside your broader biological understanding to assess the causal relationship.}\newline
Q: Please quantify the extent \ \dots \\
Literature evidence & \dots \ follows instructions and returns only 2dp probabilities as answers.\newline
\textbf{You may use the following contextual information, which are extracted from research articles:\newline
We showed that treatment of tumor cells with a DNA-damage response (DDR) inhibitor targeting the ATR kinase limits the induction of CD47 and PD-L1 signals thus promoting increased antitumor abscopal activity in vivo. \newline
ATR inhibition may cause downregulation of programmed cell death 1 ligand 1 (PD-L1) and leukocyte surface antigen 47 (CD47), thereby giving a partial suppression of the PD-1/PD-L1 and SIRP$\alpha$/CD47 immune checkpoints. }\newline
Q: Please quantify the extent \ \dots \\
Literature evidence as supplementary information & \dots \ follows instructions and returns only 2dp probabilities as answers.\newline
\textbf{You may use the following contextual information, extracted from research articles by PubTator. These provide associations or findings reported in the literature and are useful as supplementary evidence. However, they are not comprehensive or definitive sources for causal relationships. Consider the possibility of differences in biological contexts, experimental setups, or interpretation. Use them alongside your broader understanding of gene interactions, regulatory pathways, and experimental evidence.\newline
Literature evidence as supplementary information:\newline
We showed that treatment, \ \dots \ immune checkpoints. \newline
Note: The above associations from PubTator provide context about findings in specific studies. They may reflect particular experimental conditions or biological contexts that are not universally applicable. While they may suggest plausible evidence in favour or against a causal relationship, they should not be treated as the sole or definitive source of information. Always integrate this context with your broader biological knowledge and reasoning to assess the causal relationship accurately.}\newline
Q: Please quantify the extent \ \dots \\ \bottomrule
\end{tabular}
\label{tab:prompt_details_gene_specific}
\end{table}

\begin{table}[ht]
\caption{Prompts with chain-of-thought}
\centering
\small
\renewcommand{\arraystretch}{1.5} 
\begin{tabular}{p{3cm}p{11cm}} 
\textbf{Chain of thought \newline variant} & \textbf{Prompt addition} \\ \toprule
Simple & \dots \ Return your answer as a 2 decimal place probability between 0 and 1. \textbf{Please think step by step. Remember to first give one or two sentences of reasoning, and then YOU MUST include a 2dp probability at the end of your answer!}\newline
LLM output:\newline
\textbf{Reason = \{Here you will include reasoning\}}\newline
\dots \ Return your answer as a 2 decimal place probability between 0 and 1. \textbf{Please think step by step. Remember to first give one or two sentences of reasoning, and then YOU MUST include a 2dp probability at the end of your answer! \newline
Reason =}\\
Guided & \dots \ Return your answer as a 2 decimal place probability between 0 and 1. \textbf{Please think step by step. To do so first, generate evidence and reasoning in favour of a causal effect. Second, generate evidence and reasoning against a causal effect. Third, come to an overall conclusion about the evidence. Fourth, include a 2dp probability that reflects, overall, the extent to which [Gene A] has a causal effect on [Gene B]. YOU MUST include a 2dp probability at the end of your answer! }\newline
LLM output: \newline
\textbf{Evidence/reasoning in favour of causal effect = \{Relevant evidence/reasoning in favour\}\newline
Evidence/reasoning against a causal effect = \{Relevant evidence/reasoning against\}\newline
Overall conclusion = \{A simple summary of the evidence/reasoning\}}\newline
Probability = \{the 2dp probability\}\newline
\dots Return your answer as a 2 decimal place probability between 0 and 1. \textbf{Please think step by step. To do so first, generate evidence and reasoning in favour of a causal effect. Second, generate evidence and reasoning against a causal effect. Third, come to an overall conclusion about the evidence. Fourth, include a 2dp probability that reflects, overall, the extent to which ATR has a causal effect on CD47. YOU MUST include a 2dp probability at the end of your answer!\newline
Evidence}
\end{tabular}
\label{tab:prompt_details_CoT}
\end{table}

\begin{table}[ht!]
\caption{Contingency table showing the relationship between STRINGdb-derived literature gene--gene associations and whether the corresponding edge is causal.}
\centering
\begin{tabular}{lcc}
\\
\toprule
\textbf{} & \textbf{No literature evidence} & \textbf{Literature evidence present} \\ 
\midrule
\textbf{No causal edge} & 4060           & 5639           \\ 
\textbf{Causal edge}     & 78           & 123            \\ 
\bottomrule
\end{tabular}
\label{tab:pubtator_fishers}
\end{table}

\begin{figure}[!ht]
\centering
\includegraphics[width=\textwidth]{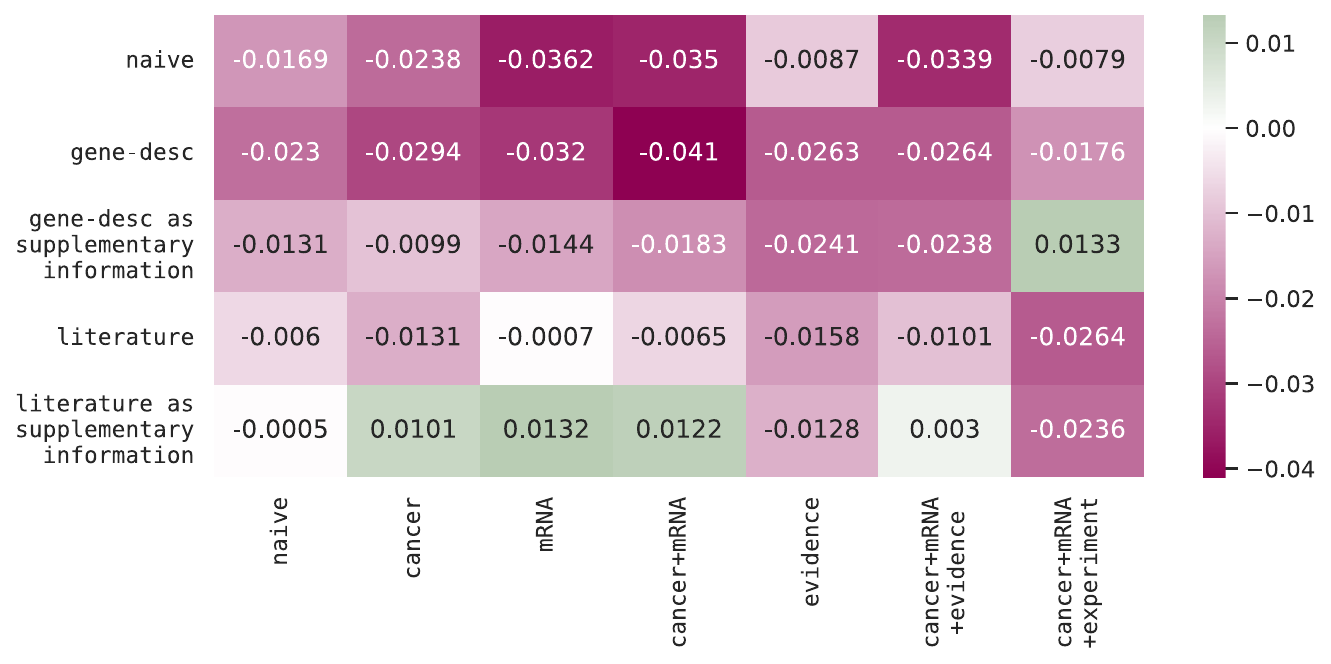}
\caption{Results on Gemma2 using simple chain of thought, compared to the results using no chain of thought (shown in Figure \ref{fig:gemma2_results_matrix}). Green indicates the simple CoT reached a higher AUROC than no CoT and pink indicates it reached a lower AUROC than no CoT.}
\label{fig:gemma2_results_cot_simple}
\end{figure}

\begin{figure}[!ht]
\centering
\includegraphics[width=\textwidth]{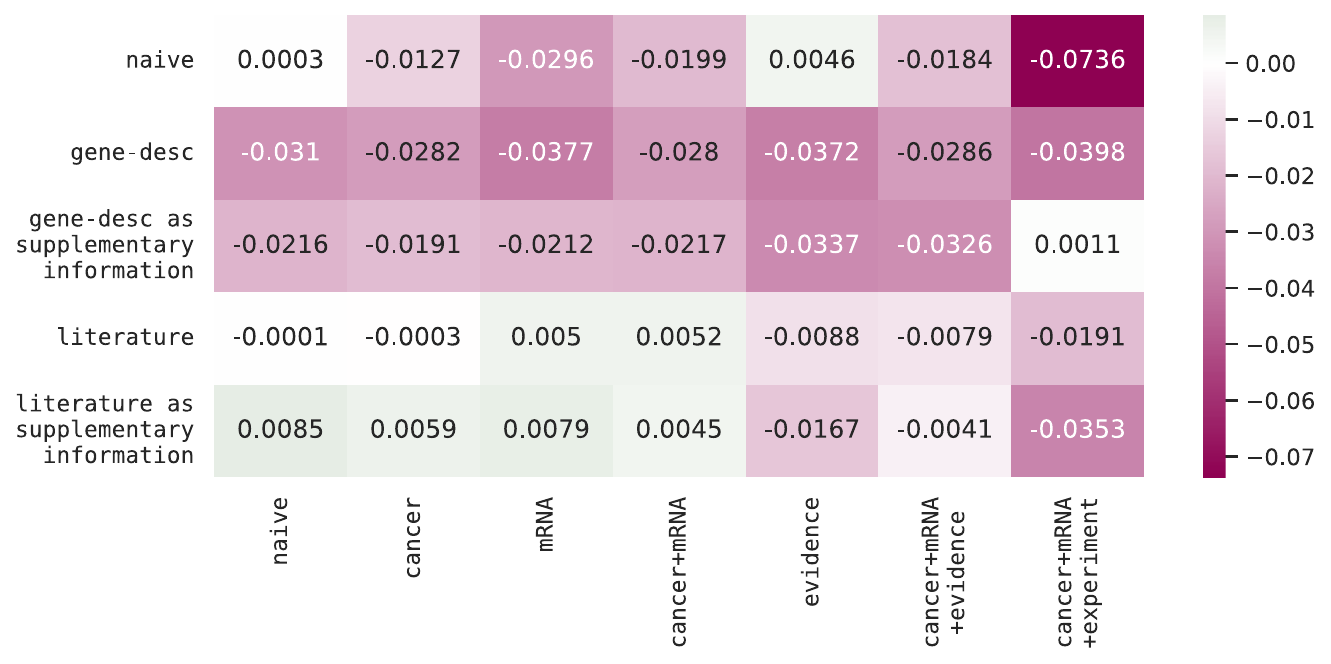}
\caption{Results on Gemma2 using guided chain of thought, compared to the results using no chain of thought (shown in Figure \ref{fig:gemma2_results_matrix}). Green indicates the guided CoT reached a higher AUROC than no CoT and pink indicates it reached a lower AUROC than no CoT.}
\label{fig:gemma2_results_cot_guided}
\end{figure}

\begin{figure}[!ht]
\centering
\includegraphics[width=\textwidth]{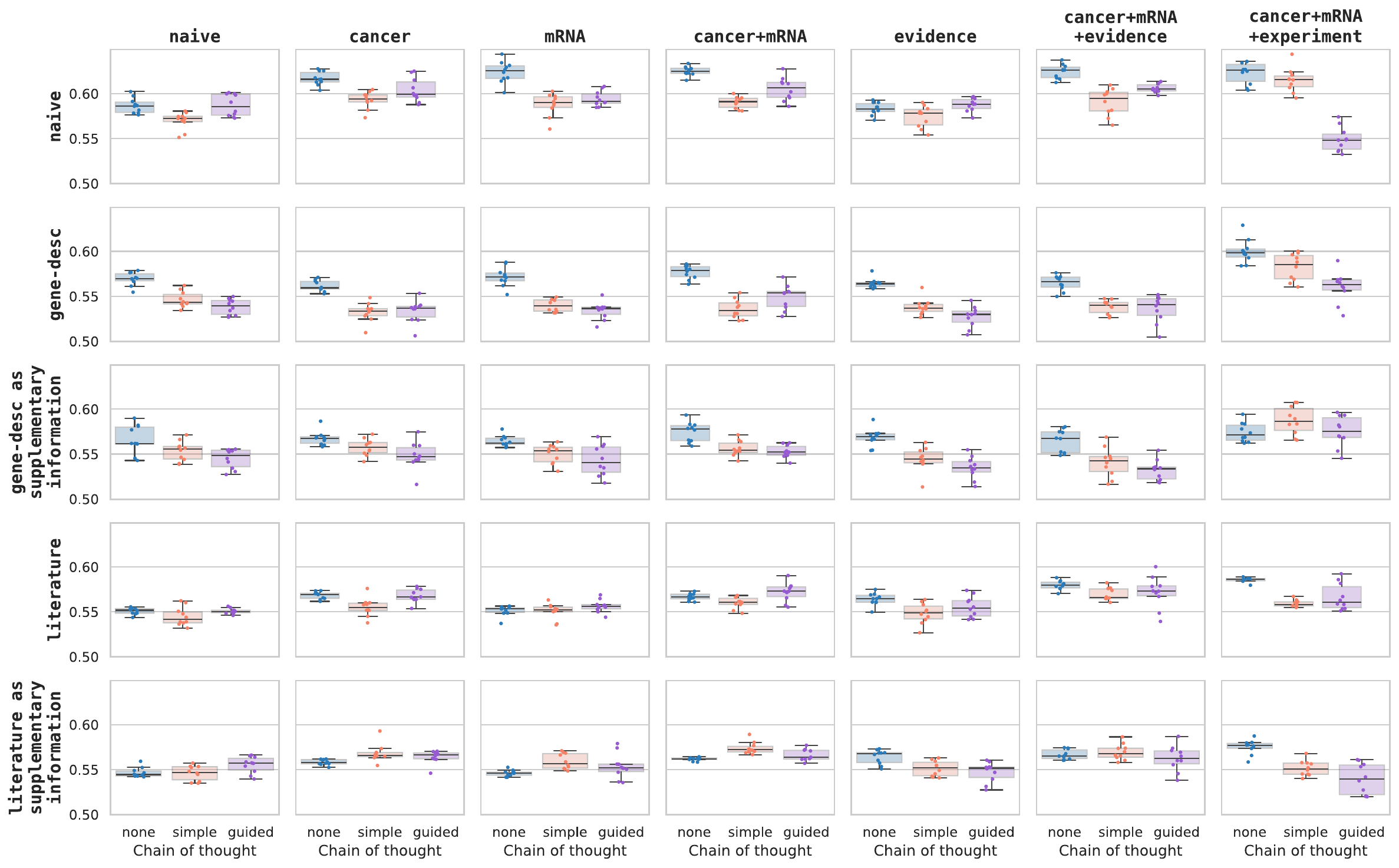}
\caption{Results on Gemma2 for all combinations of prompt variants (different contexts along each column, different gene-specific information along each row), shown as boxplots. The AUROCs for each of the 10 repetitions are plotted on the y axes.}
\label{fig:gemma2_results_boxplots}
\end{figure}


\end{document}